\pdfoutput=1

\documentclass[11pt]{article}

\usepackage{EMNLP2022}

\usepackage{times}
\usepackage{latexsym}

\usepackage[T1]{fontenc}

\usepackage[utf8]{inputenc}

\usepackage{microtype}

\usepackage{inconsolata}

\usepackage{amsmath}
\usepackage{amsfonts}
\usepackage{subcaption}
\usepackage{graphicx}
\usepackage{booktabs}
\usepackage{url}
\usepackage{tabularx}
\usepackage{bigstrut}
\usepackage{multirow}
\usepackage{color}
\usepackage{xcolor}
\usepackage{verbatimbox}
\usepackage{enumitem}
\usepackage{xspace}

\def\ie{\emph{i.e}.}
\def\eg{\emph{e.g}.}
\def\ie{\emph{i.e}.}

\def\etc{\emph{etc}}

\newcommand{\MODEL}{MGDoc\xspace}

\title{MGDoc: Pre-training with Multi-granular Hierarchy\\for Document Image Understanding}

\author{
Zilong Wang$^{1}$\thanks{\hspace{.1in}This work was completed while the author was working as an intern at Adobe Research.} \quad Jiuxiang Gu$^{2}$ \quad Chris Tensmeyer$^{2}$ \quad Nikolaos Barmpalios$^{2}$ \\
\textbf{Ani Nenkova}$^{2}$ \quad \textbf{Tong Sun}$^{2}$ \quad \textbf{Jingbo Shang$^{1}$} \quad \textbf{Vlad I. Morariu}$^{2}$ \\
$^1$University of California, San Diego \qquad $^2$Adobe Research \\
$^1$\texttt{\{zlwang, jshang\}@ucsd.edu} \\
$^2$\texttt{\{jigu, tensmeye, barmpali, nenkova, tsun, morariu\}@adobe.com}
}

\begin{document}
\maketitle
\begin{abstract}
Document images are a ubiquitous source of data where the text is organized in a complex hierarchical structure ranging from fine granularity (e.g., words), medium granularity (e.g., regions such as paragraphs or figures), to coarse granularity (e.g., the whole page). The spatial hierarchical relationships between content at different levels of granularity are crucial for document image understanding tasks. Existing methods learn features from either word-level or region-level but fail to consider both simultaneously. Word-level models are restricted by the fact that they originate from pure-text language models, which only encode the word-level context. In contrast, region-level models attempt to encode regions corresponding to paragraphs or text blocks into a single embedding, but they perform worse with additional word-level features. To deal with these issues, we propose MGDoc, a new multi-modal multi-granular pre-training framework that encodes page-level, region-level, and word-level information at the same time. MGDoc uses a unified text-visual encoder to obtain multi-modal features across different granularities, which makes it possible to project the multi-granular features into the same hyperspace. To model the region-word correlation, we design a cross-granular attention mechanism and specific pre-training tasks for our model to reinforce the model of learning the hierarchy between regions and words. Experiments demonstrate that our proposed model can learn better features that perform well across granularities and lead to improvements in downstream tasks.

\end{abstract}

\section{Introduction}

\begin{figure}[t]
    \centering
    \includegraphics[width=\linewidth]{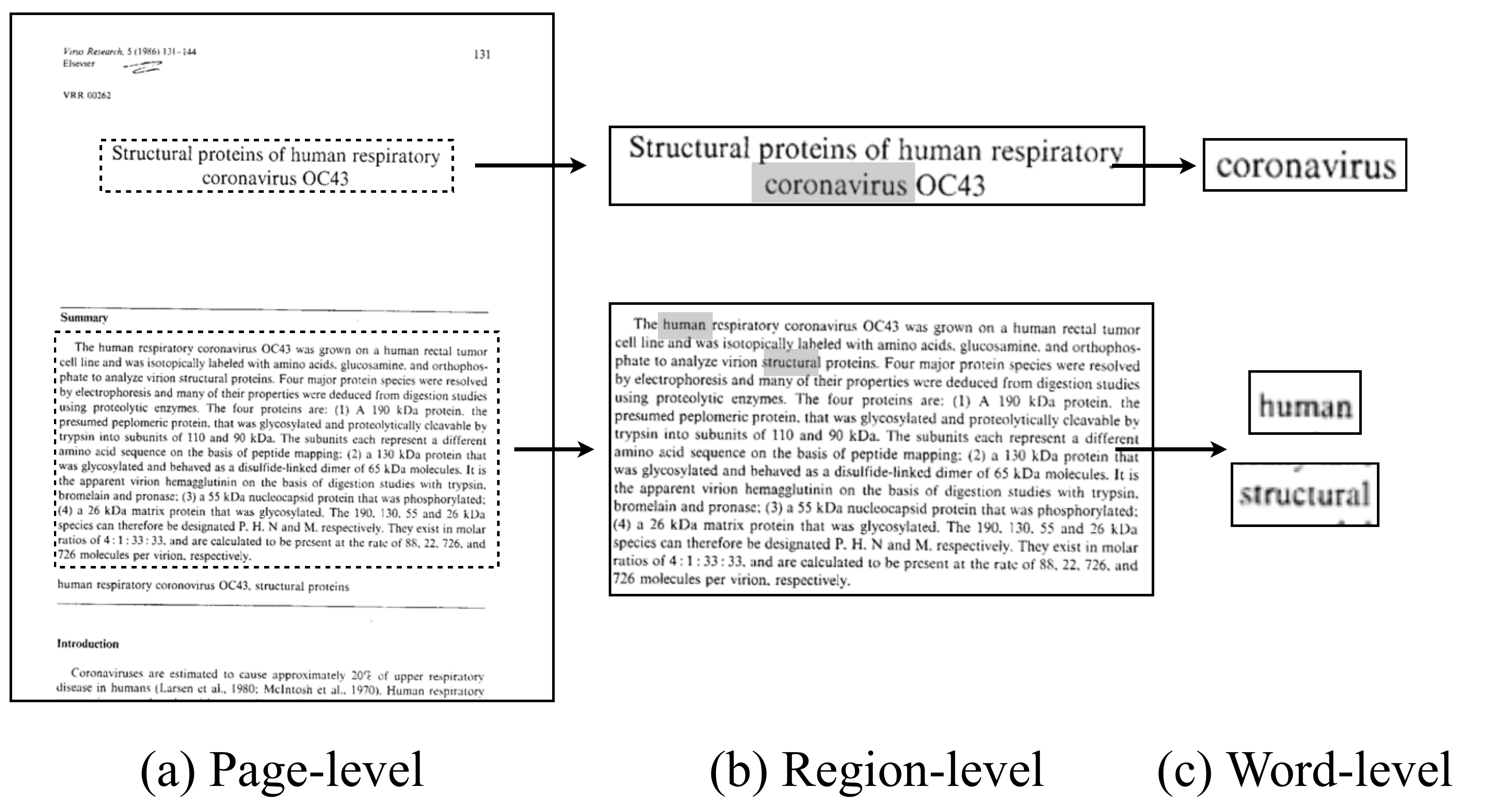}
    \caption{The multi-granular structure in document images. The image is from the  RVL-CDIP dataset. Important information is encoded at the page-level (\eg, the document type), region-level (constructs such as paragraphs, tables,~\etc), and word-level (specific semantics). Our proposed model reasons about all of these jointly to leverage information across granularities.}
    \label{fig:mg-example}
\end{figure}

Document images are ubiquitous and are often used as a representation for forms, receipts, printed papers, etc. Unlike plain text documents, document images express rich information via both textual content and heterogeneous layout patterns, which leads to barriers to the automatic processing of these document images. Here, \textit{layout pattern} refers to how the text is spatially arranged on the document page and involves information from multiple levels of granularity. Specifically, a layout pattern divides the entire page into individual regions and, within each region, the fine-grained textual content is distributed following a certain format, such as paragraphs, columns, lists, as shown in Figure \ref{fig:mg-example}. 

The layout of a document provides important cues for interpreting the document through spatial structures such as alignment, proximity, and hierarchy between content at different levels of granularity. For example, a numeric text field is more likely to be the total price of a grocery receipt if it is located at the bottom right of a table region; a region is more likely to correspond to the title area of a form if there are a lot of bold types inside of the region. In these two examples, it is important to understand page-level information (\eg, that the document is a receipt or a form), region-level information (\eg, that a region is a table or title), and word/token-level information (\eg, the font style of a word, or that a token is a number), as well as how these relate to each other. Therefore, to facilitate the automatic processing of such documents, it is essential to consider the features of multiple granularities and let the model learn the hierarchy between different levels to encode the multi-granular structure in the document images. 


\begin{figure}[t]
    \centering
    \includegraphics[width=0.8\linewidth]{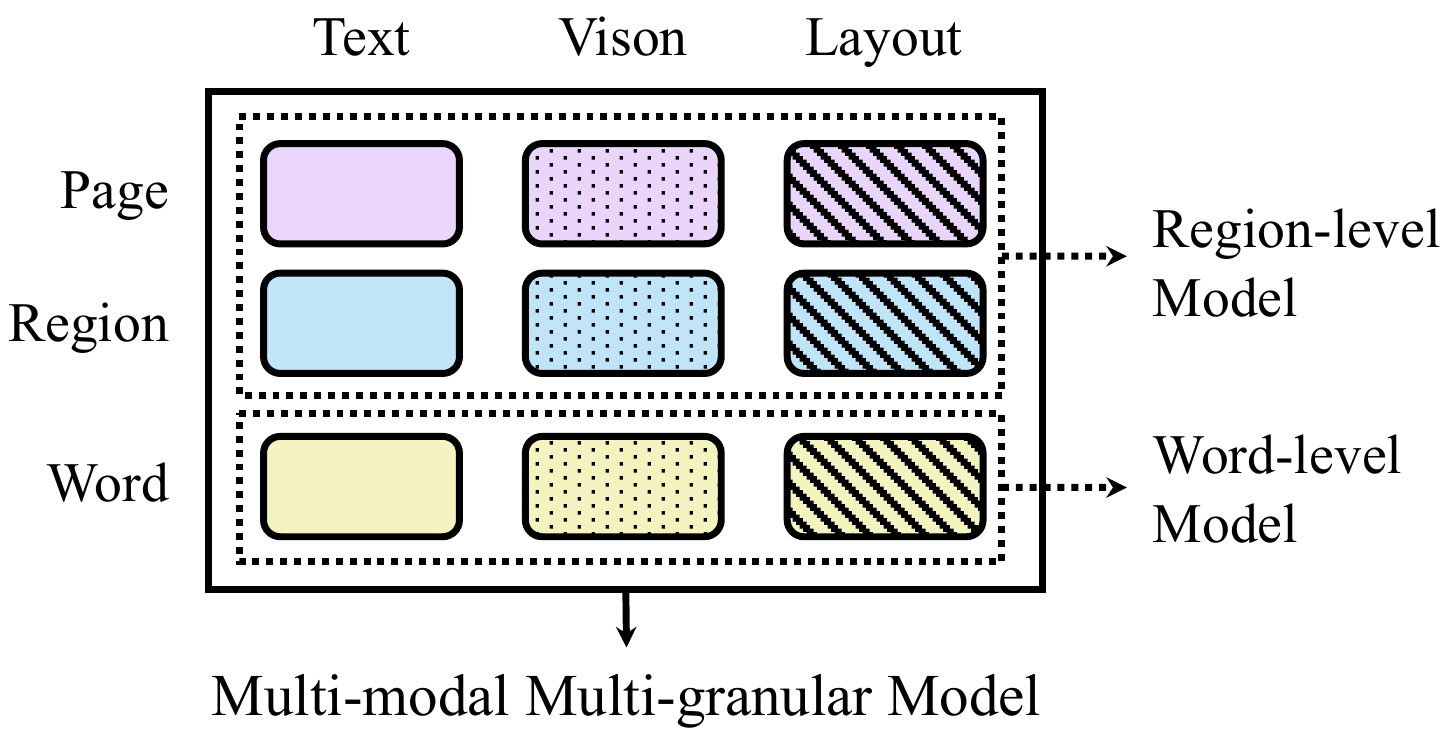}
    \caption{Comparison between \MODEL and existing methods. While previous methods have explored multi-modal inputs, and explored word+page and region+page level features, MGDoc combines multi-modal reasoning, and joint word-, region-, and page-level reasoning.}
    \label{fig:mm-mg}
\end{figure}

However, existing methods in document image understanding formulate the document image understanding tasks either at the word-level or region-level and thus do not use both cues. 
They mostly follow language modeling methods designed for plain text settings, formulating document image understanding tasks using word-level information and augmenting semantic features with spatial and visual features to exploit the word-level context~\cite{xu2020layoutlm,xu2020layoutlmv2,hong2021bros,garncarek2021lambert}. 
Recent works go beyond fine-grained word-level inputs and focus on regions instead of words to acquire useful signals~\cite{li2021selfdoc,gu2021unidoc}. By encoding the regions corresponding to paragraphs or text blocks, these region-level models manage to save training resources and achieve good performance with rich locality features. However, these models fail to leverage the cross-region word-level correlation, which is also necessary to tackle fine-grained tasks.

Motivated by this observation, we propose \MODEL, a new multi-modal multi-granular pre-training framework that encodes document information at different levels of granularity and represents them using multi-modal features as highlighted in Figure \ref{fig:mm-mg}.
Specifically, we use the OCR engine to decompose a document page into three granularities: page-level, region-level, and word-level. Following previous works~\cite{xu2020layoutlm,xu2020layoutlmv2,gu2021unidoc}, our multi-modal features represent text, layout (represented by bounding boxes), and image modalities. The input consists of information at different levels of granularity, and can be organized into a hierarchy within the page, which means words are included in the corresponding regions and the page includes all of them. We leverage attention to learn the correlation between inputs from different levels of granularity and add special attention weights to encode the hierarchical structure and relative distances~\cite{xu2020layoutlmv2,garncarek2021lambert,powalski2021going}. We rely on pre-training to encourage the model to learn the alignment between regions at different levels of granularity. 
In addition, we use masked language modeling for the word-level inputs and extend this idea into the more coarse-grained inputs. We mask a proportion of regions and ask the model to minimize the difference between the masked contextual features and the input features corresponding to the selected region.

We validate \MODEL on three public benchmarks, the FUNSD dataset~\cite{jaume2019funsd} for form understanding, the CORD dataset~\cite{park2019cord} for receipt extraction, and the RVL-CDIP dataset~\cite{lewis2006building} for document image classification. Extensive experiments demonstrate the effectiveness of our proposed approach with great improvements on fine-grained tasks and good results on coarse-grained tasks. We summarize our contribution as follows:
\begin{itemize}[leftmargin=*,nosep]
\item We propose \MODEL, a multi-modal multi-granular pre-training framework, which encodes the hierarchy in document images and integrates features from text, layout patterns, and images.
\item A cross-granularity attention mechanism and a new pre-training task designed to enable the model to learn the alignment between different levels. This work extends the masked language modeling to different granularity to encode the contextual information.
\item Extensive experiments demonstrate the effectiveness of \MODEL on three representative benchmarks.

\end{itemize}

\begin{figure*}[t]
    \centering
    \includegraphics[width=\linewidth]{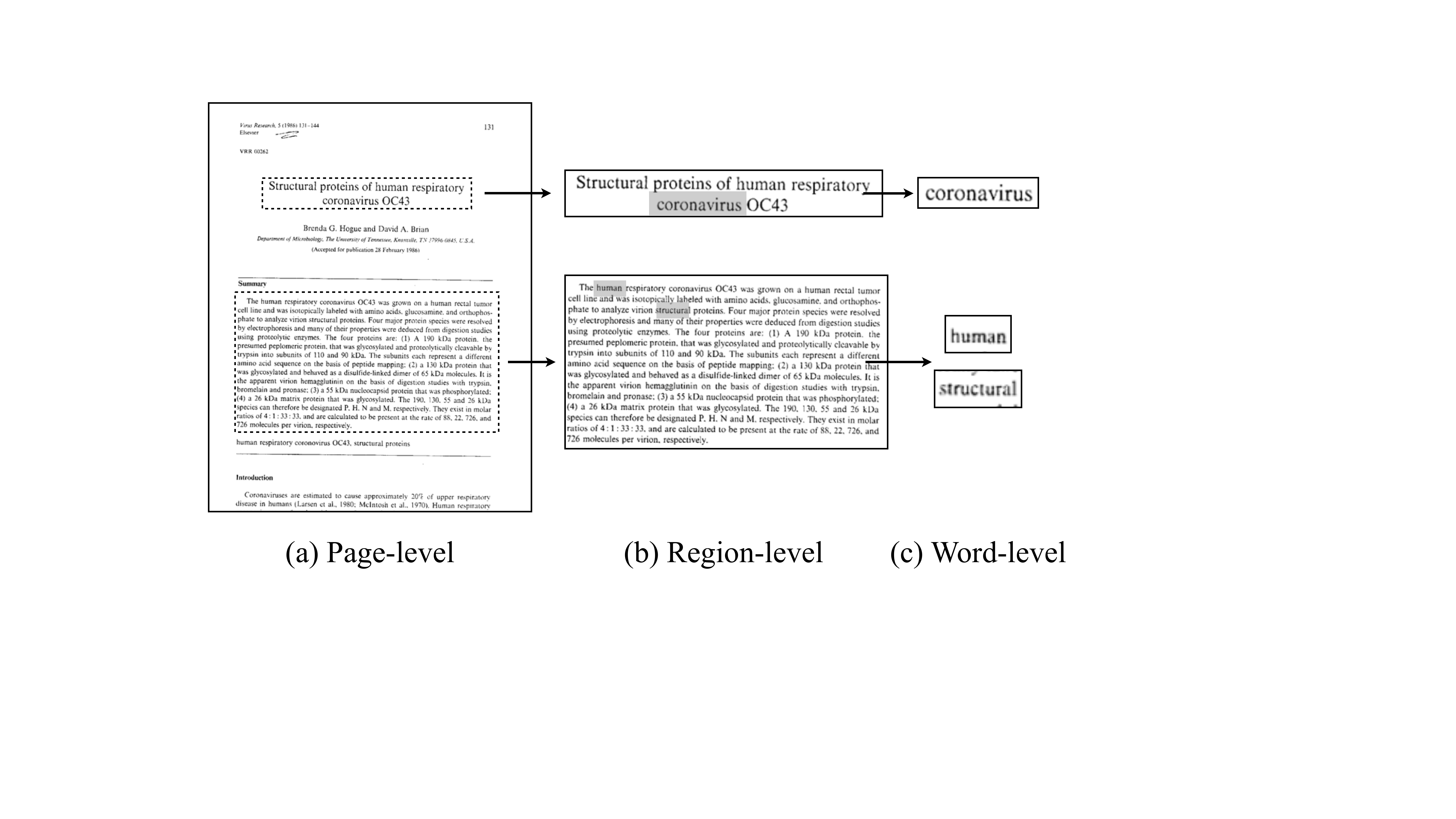}
    \caption{The multi-modal multi-granular pre-training framework of \MODEL. Inputs consist of visual, textual, and spatial representations of the page, regions on the page, and individual words. Multi-granular attention learns relationships within and across granularities within single modalities, followed by multi-granular attention across modalities. The final output consists of an embedding for each region at each granularity, on top of which three self-supervision tasks are added to pre-train the model. Specifically, the multi-granularity task ensures that our model makes use of multi-granular inputs and multi-granular attention to solve spatial tasks.}
    \label{fig:model}
\end{figure*}

\section{Method}

\subsection{Overview}
\MODEL is a multi-modal multi-granular pre-training framework for document image understanding tasks. The framework encodes features from different levels of granularity in a document page and leverages the spatial hierarchy between them. There are three stages in our architecture. First, OCR engine, human annotations, or digital parsing provide us with the text and the bounding boxes of contents at different levels of granularity. We focus on pages, regions, and words in this paper and leave the more fine-grained pixel-level and coarse-grained multi-page modeling for future work. 
We input the document image, textual content, and bounding boxes at these three levels to the multi-modal encoder to encode multi-granular information into text and image embeddings. Following previous work~\cite{xu2020layoutlm}, we use spatial embeddings to encode spatial layout information. Next, we design a multi-granular attention mechanism to extract the correlation between the features from different levels. Distinct from the normal self-attention mechanism in BERT~\cite{lu2019vilbert}, multi-granular attention computes the dot product between the features and encodes the hierarchical relationship between regions and words by adding an attention bias. Then, the cross-attention mechanism is used to combine the features from different modalities. Finally, the sum of the final text and visual features is used in the pre-training or fine-tuning tasks.

\subsection{Multi-modal Encoding}\label{sec:mm-encoding}
The multi-modal encoding is designed to encode the text, visual, and spatial information of the multi-granular inputs into the embedding space. We first acquire the inputs at the word-level, encoding the text and bounding box of each word. At the region-level, words are grouped into regions where all words within the region are combined and the enclosing bounding box of all the words is used as the region bounding box. At the page-level, the textual input is the sequence of all words in the page and the width and height of the document image is used as the page-level bounding box. Now, the inputs at different levels of granularity consist of the textual content and the bounding box.
We denote the inputs as $P$, $\{R1,...,R_m\}$, $\{w_1,...,w_n\}$, where $m,n$ are the number of regions and words. 


The multi-modal encoding takes the textual content of each input unit, ranging from a single word, to sentences, to the whole textual content of a page, and encodes it with a pre-trained language model,~\eg, SBERT~\cite{reimers2019sentence}. We add the spatial embeddings to the text encoder outputs where the fully-connected layer is used to project the bounding boxes into a hyperspace. In this way, the text embeddings of our model are augmented with spatial information.
Then, \MODEL encodes the entire image with a visual feature extractor,~\eg, ResNet~\cite{he2016deep}, and extracts region feature maps using bounding boxes as the Region of Interest (ROI) areas. The results of the vision encoder have different resolutions due to the sizes of bounding boxes, but they lie in the same feature space. Similarly, we add the spatial embeddings from the bounding boxes to the visual embeddings.
The multi-modal embeddings are represented as follows,
\begin{align*}
    e^T_\lambda &= \text{Enc}^T(text_{\lambda}) + \text{FC}(box_{\lambda}) + \text{Emb}^T \\
    e^V_\lambda &= \text{Enc}^V(img)[box_\lambda] + \text{FC}(box_{\lambda}) + \text{Emb}^V 
\end{align*}
where $e^T_\lambda$ and $e^V_\lambda$ are the text and visual embeddings; $\lambda\in \{P\}\cup\{R_1,...,R_m\}\cup\{w_1,...,w_n\}$ denotes different levels of granularity; $\text{Enc}^T$ and $\text{Enc}^V$ are the text and image encoders, respectively; $text$ and $box$ refer to the textual contents and bounding boxes; $img$ is the entire document image; FC$(\cdot)$ is the fully-connected layer; $\text{Emb}^T$ and $\text{Emb}^V$ are the type embeddings for text and vision.

\subsection{Multi-granular Attention}\label{sec:multi-granular}
Given the multi-modal embeddings described above, we design a multi-granular attention to encode the hierarchical relation between regions and words. Specifically, we add attention biases to the original self-attention weights to strengthen the region-word correlation. We apply multi-granular attention to the text embeddings and visual embeddings individually because the purpose of this module is to learn the interaction between different levels of granularity rather than to fuse modalities. Therefore, without loss of generality, we omit the notation of modality in the expressions. The attention weight is computed as
\begin{align*}
    A_{\alpha,\beta} &= \frac{1}{\sqrt{d}}(W^Q e_\alpha)^\top(W^K e_\beta) \\
                     &+ \text{HierBias}(box_{\alpha}\subseteq \text{or} \not\subseteq box_{\beta})\\
                     &+ \text{RelBias}(box_{\alpha}- box_{\beta}),
\end{align*}
where $\alpha,\beta \in \{P\}\cup\{R_1,...,R_m\}\cup\{w_1,...,w_n\}$; the first part is the same as the attention mechanism in the original BERT; $\text{RelBias}(\cdot)$ is the attention weight bias to encode the relative distance between the bounding boxes; $\text{HierBias}(\cdot)$ is the attention weight bias to encode the inside or outside relation which models the spatial hierarchy within the page. Since the regions are created by grouping the words, all the words correspond to a specific region. We embed the binary relation into a fixed-sized vector and each value in this vector is added to an attention head of the multi-granular attention module.
After the multi-granular attention module, self-attention is applied to the input embeddings to learn contextual information and hierarchical relationships between the multi-granular inputs. We denote the resulting  textual and visual features by $f^T_\lambda$ and $f^V_\lambda$, respectively, where $\lambda\in \{P\}\cup\{R_1,...,R_m\}\cup\{w_1,...,w_n\}$.

\subsection{Cross-modal Attention}
As mentioned in Section \ref{sec:multi-granular}, multi-granular attention is only designed for the interaction between different levels of granularity; however, it is also essential to fuse information from multiple modalities. Therefore, we design the cross-modal attention module to conduct the modality fusion. Following previous works in visual-language modeling, we use a cross-attention mechanism to fuse the textual and visual features. Specifically, the cross-attention function is formulated as,
\begin{align*}
    \text{CrossAttn}(\alpha|\beta) &= \sigma \left(\frac{(W^Q \alpha)^\top(W^K \beta)}{\sqrt{d}}\right)W^V\beta
\end{align*}
where $\alpha$ and $\beta$ are matrices of the same size; $\sigma$ is the Softmax function to normalize the attention matrix; $W^Q,W^K,W^V$ are the trainable weights in each of the attention heads for query, key and, value.
Then we list text and visual features from different levels of granularity as $F^T=\{f^T_\lambda\}$ and $F^V=\{f^V_\lambda\}$ and compute the multi-modal features as follows,
\begin{align*}
    f^{T\to V}_\lambda &= \text{CrossAttn}(F^T|F^V)[\lambda] \\
    f^{V\to T}_\lambda &= \text{CrossAttn}(F^V|F^T)[\lambda]
\end{align*}

From these expressions, we can see that the cross attention uses the dot product between multi-modal features as attention weights. In this way, the given modality can learn from the other modality and the module also bridges the gap between the modalities. We call the output of this module text or visual multi-modal features to distinguish from the text and visual features in Section \ref{sec:mm-encoding}, and denote them as $f^{T\to V}_\lambda$ and $f^{V\to T}_\lambda$.
The final representation is the sum of the textual and visual multi-modal features, $f_\lambda=f^{T\to V}_\lambda+f^{V\to T}_\lambda, \lambda\in \{P\}\cup\{R_1,...,R_m\}\cup\{w_1,...,w_n\}$, and is used in the pre-training and downstream tasks.

\subsection{Pre-training Tasks}
Large-scale pre-training has shown strong results in document image understanding tasks. With a large amount of unlabeled data, pre-trained models can learn the latent data distribution without manually labelled supervision and can easily transfer the learned knowledge to downstream tasks. The design of pre-training tasks is crucial for successful pre-training. We go beyond the classic mask modeling and apply the mask text modeling and the mask vision modeling on all the inputs from different levels of granularity. Due to the unified multi-model encoder (see Section \ref{sec:mm-encoding}), it is possible for us to treat all levels of granularity equally and introduce a unified masking task for each modality. Because we believe that spatial hierarchical relationships are essential for encoding documents, we design a pre-training task that requires the model to identify the spatial relationship between content at different levels of granularity. The final training loss is the sum of the pre-training tasks, $\mathcal{L} = \mathcal{L}_{MTM} +  \mathcal{L}_{MVM}+\mathcal{L}_{MGM}$.
Below we provide details for each component.

\paragraph{Mask Text Modeling}
The mask text modeling task requires the model to understand the textual inputs of the model. Specifically, we randomly select a proportion of regions or words, and their textual contents are replaced with a special token \texttt{[MASK]}. We run the model to obtain the contextual features of these masked inputs and compare them with the encoding result of original textual inputs. We use the Mean Absolute Error as the loss function. 
\begin{align*}
    \mathcal{L}_{MTM} = \sum_{\lambda\in \Lambda} \left|e^T_{\lambda}-f^{V\to T}_{\texttt{[MASK]}|\Bar{\lambda}}\right|
\end{align*}
where $\Lambda = \{P\}\cup\{R_1,...,R_m\}\cup\{w_1,...,w_n\}$; $e^T_\lambda$ denotes the encoding result of the original textual contents; $\Bar{\lambda}$ denotes the multi-granular context without $\lambda$; $f^{V\to T}_{\texttt{[MASK]}|\Bar{\lambda}}$ is the contextual feature of the masked textual inputs.

\paragraph{Mask Vision Modeling}
Similarly to the mask text modeling task, we use mask vision modeling to learn visual contextual information. Instead of replacing the \texttt{[MASK]} token as is done in mask text modeling,  we set the visual embeddings of the selected areas to zero vectors. The loss function computes the Mean Absolute Error between the contextual feature of masked areas and the original visual embeddings. The mask vision modeling loss is formulated as,
\begin{align*}
    \mathcal{L}_{MVM} = \sum_{\lambda\in \Lambda} \left|e^V_{\lambda}-f^{T\to V}_{\texttt{[0]}|\Bar{\lambda}}\right|
\end{align*}
where $f^{T\to V}_{\texttt{[0]}|\Bar{\lambda}}$ is the contextual feature of the zero vector given the unmasked inputs.

\paragraph{Multi-Granularity Modeling}
The multi-granularity modeling task asks the model to understand the spatial hierarchy between different levels of granularity. Since the page-level input includes all regions and words, it is trivial for the model to learn it. We only focus on the hierarchical relation between the regions and words. Although the relation is also encoded in the multi-granular attention, it is necessary to reinforce the model to emphasize the region-word correspondence. Otherwise, the spatial hierarchy biases are random add-ons to the attention matrix. 

The model takes the region-level and word-level features and predicts which region the given the word is located in. We first compute the dot product of the region-level and word-level features as the score and use the Cross-entropy as the loss function.
\begin{align*}
    \mathcal{L}_{MGM} = \sum_{w\in W} \frac{e^{f_w^\top f_{r^*}}}{e^{f_w^\top f_{r^*}} + \sum_{r\in R-\{r^*\}} e^{f_w^\top f_{r}}}
\end{align*}
where $W = \{w_1,...,w_n\}$ and $R = \{R_1,...,R_m\}$; $r^*$ is the region that includes the word $w$.

\section{Experiments}

\subsection{Pre-training Settings}
We use the RVL-CDIP dataset~\cite{harley2015icdar} as our pre-training corpus. The RVL-CDIP dataset is a scanned document image dataset containing 400,000 grey-scale images and covering a variety of layout patterns. We use OCR engines to recognize the location of textual content in the document images and also the location of the individual words. Following \citet{gu2021unidoc}, we use EasyOCR~\footnote{\url{https://github.com/JaidedAI/EasyOCR}} with two different output modes: non-paragraph and paragraph. The difference is that the non-paragraph mode extracts the individual words in the pages, and the paragraph mode groups these results into regions. The OCR engine allows us to design the architecture and the pre-training tasks focusing on the multi-granularity of document images. Therefore, the paragraph results serve as the region-level inputs, and the non-paragraph results serve as the word-level inputs.

\begin{table*}[t]
  \centering
  \small
\resizebox{\linewidth}{!}{
    \setlength{\tabcolsep}{4mm}{
    \begin{tabular}{clcccccc}
    \toprule
    \multirow{2}[4]{*}{\textbf{Scale}} & \multirow{2}[4]{*}{\textbf{Model}} & \multicolumn{3}{c}{\textbf{Pre-training}} & \textbf{FUNSD} & \textbf{CORD} & \textbf{RVL-CDIP} \\
\cmidrule{3-5}          &       & \textbf{Corpus} & \textbf{\#Data} & \textbf{\#Param.} & \textbf{(F1)} & \textbf{(F1)} & \textbf{(Accuracy)} \\
    \midrule
    \multirow{12}[2]{*}{Word} & BERT$_{\text{BASE}}$ & -    & -    & 110M  & 60.26 & 89.68 & 89.81 \\
          & BERT$_{\text{LARGE}}$ & -    & -    & 340M  & 65.63 & 90.25 & 89.92 \\
          & LayoutLM$_{\text{BASE}}$ & IIT-CDIP & 11M   & 113M  & 78.66 & 94.72 & 94.42 \\
          & LayoutLM$_{\text{LARGE}}$ & IIT-CDIP & 11M   & 343M  & 78.95 & 94.93 & 94.43 \\
          & BROS$_{\text{BASE}}$ & IIT-CDIP & 11M   & 110M  & 83.05 & 96.50 & - \\
          & BROS$_{\text{LARGE}}$ & IIT-CDIP & 11M   & 340M  & 84.52 & 97.28  & - \\
          & LayoutLMv2$_{\text{BASE}}$ & IIT-CDIP & 11M   & 200M  & 82.76 & 94.95 & 95.25 \\
          & LayoutLMv2$_{\text{LARGE}}$ & IIT-CDIP & 11M   & 426M  & 84.20  & 96.01 & 95.64 \\
          & TILT$_{\text{BASE}}$ & RVL-CDIP+ & 1.1M  & 230M  & -    & 95.11 & 95.25 \\
          & TILT$_{\text{LARGE}}$ & RVL-CDIP+ & 1.1M  & 780M  & -    & 96.33 & 95.52 \\
          & DocFormer$_{\text{BASE}}$ & IIT-CDIP- & 5M    & 183M  & 83.34 & 96.33 & 96.17 \\
          & DocFormer$_{\text{LARGE}}$ & IIT-CDIP- & 5M    & 536M  & 84.55 & 96.99 & 95.50 \\
    \midrule
    \multirow{4}[2]{*}{Region} & SelfDoc & RVL-CDIP & 320K  & -    & 83.36 & -    & 92.81 \\
          & SelfDoc+VGG-16 & RVL-CDIP & 320K  & -    & -    & -    & 93.81 \\
          & UDoc  & IIT-CDIP- & 1M    & 272M  & 87.96 & 96.64 & 93.96 \\
          & UDoc$\ddag$  & IIT-CDIP- & 1M    & 272M  & 87.93 & 96.86 & 95.05 \\
    \midrule
    \textbf{Region+Word} & \textbf{MGDoc (Ours)} & \textbf{RVL-CDIP} & \textbf{320K}  & \textbf{203M}$^*$  & \textbf{89.44} &  \textbf{97.11} & \textbf{93.64} \\
    \bottomrule
    \end{tabular}%
    }} 
      \caption{The experiment results and comparison. * indicates that non-trainable parameters are not included. The total \#param. of \MODEL is 312M. $\ddag$ implies unfreezing the sentence encoder during the finetuning. RVL-CDIP+: TILT uses extra training pages in pre-training; IIT-CDIP-: DocFormer uses a subset of IIT-CDIP in pre-training.}
  \label{tab:result}%
\end{table*}%

\subsection{Fine-tuning Tasks}
We select three representative tasks to evaluate the performance of our model and use the publicly-available benchmarks for each tasks.

\paragraph{Form Understanding}
The goal of the form understanding task is to predict the label of semantic entities in document images. We use the FUNSD dataset~\cite{jaume2019funsd} for this task. The FUNSD dataset consists of 199 fully-annotated, noisy-scanned forms with various appearances and formats. There are 149 and 50 pages in the training set and the testing set, respectively. Each entity is labeled into 3 categories: \textit{Header}, \textit{Question}, and \textit{Answer}. We use the provided OCR results from the dataset and input the textual contents and bounding boxes of entities to the model. We report the entity-level F1 score as metrics.

\paragraph{Receipt Understanding}
The goal of the receipt understanding task is to recognize the role of a series of text lines in a document. We use the CORD dataset~\cite{park2019cord} for this task. The CORD dataset is fully annotated with bounding boxes and textual contents and contains 800 and 100 pages in the training and testing sets, respectively. There are 30 entity types marked in the dataset; we report entity-level F1 score for our experiments.

\paragraph{Document Image Classification}
The document image classification task aims to classify the pages into different semantic categories. We use the RVL-CDIP dataset~\cite{harley2015icdar} for this task, which is a subset of the IIT-CDIP dataset~\cite{lewis2006building}. The RVL-CDIP dataset contains 400,000 pages, each annotated with 16 semantic categories. The input features for this dataset are extracted by the EasyOCR engine in our experiments. The RVL-CDIP dataset is also used in the pre-training, but no labeling information is involved in the pre-training tasks, so there is no concern about data leakage. In the downstream task, the RVL-CDIP dataset is divided into training, validation, and test subsets with 8:1:1 ratio. We report classification accuracy over the 16 categories for our experiments.

\subsection{Implementation Details}
In the multi-modal encoder, we use the BERT-NLI-STSb-base model as the text encoder and ResNet-50 as the vision encoder. 
In the modality fusion, we use 12 layers of cross-modal attention in \MODEL.
We set the hidden state size as 768 and the attention head number as 12.
We freeze the pre-trained weights of the multi-modal encoder and randomly initialize the remaining parameters, which are then learned during our pre-training stage. We run the pre-training for 5 epochs with 8 NVIDIA V100 32G GPUs and the AdamW optimizer. The batch size is set to 64; the learning rate is set to $10^{-6}$; the warmup is conducted in the first 20\% training steps.

\subsection{Results}

We compare \MODEL with the strong baselines in the document understanding tasks in Table \ref{tab:result}. We list out the specific settings of each model in the layout-rich pre-training, to clearly demonstrate the effectiveness of our model. All these baseline models resort to different techniques to achieve competitive results. BERT~\cite{devlin2018bert}, LayoutLM~\cite{xu2020layoutlm}, LayoutLMv2~\cite{xu2020layoutlmv2}, BROS~\cite{hong2021bros}, TILT~\cite{powalski2021going}, and DocFormer~\cite{appalaraju2021docformer} encode word-level features, and SelfDoc~\cite{li2021selfdoc}, and UDoc~\cite{gu2021unidoc} encodes region-level features.
\MODEL surpasses all the existing methods with the help of the information from all different levels of granularity, and achieves a new state-of-the-art performance in the fine-grained tasks,~\ie, the form understanding task and receipt understanding task. It also achieves promising performance on the coarse-grained task, i.e., the document image classification task. Specifically, \MODEL improves the entity-level F1 score of the FUNSD dataset by 1.48\% and improves the entity-level F1 score of the CORD dataset by 0.25\%, compared with the second-best model. We partially attribute the performance difference on the RVL-CDIP dataset to the OCR engine, since LayoutLMv2 and TILT use the Microsoft OCR and BROS uses the CLOVA OCR, and these commercial OCR engines provide more accurate results. As discussed in \citet{gu2021unidoc}, the quality of the OCR engine influences the performance of the document image classification. It is also worth mentioning that our model involves relatively smaller number of trainable parameters and also requires less pre-training data, which makes \MODEL more applicable in realistic scenarios.

The performance of the form and receipt understanding tasks is improved by region-level information. UDoc surpasses the word-scale models by large margins, and our proposed, \MODEL, even further improves the UDoc by modeling the alignment between regions and words. We conclude that the region-level information strengthens the locality of the feature extraction, and the word-level information further improves the classification results. Such connection is realized by region-word alignment, which is visualized in Section \ref{sec:alignemnt-visualization}.

\subsection{Ablation Study}
\begin{table}[t]
  \centering
  \small
  \resizebox{\linewidth}{!}{
    \setlength{\tabcolsep}{4mm}{
    \begin{tabular}{lcc}
    \toprule
    \multirow{2}[2]{*}{\textbf{Model}} & \textbf{FUNSD} & \textbf{RVL-CDIP} \\
          & \textbf{(F1)} & \textbf{(Acurracy)} \\
    \midrule
    \MODEL \\
    \ \textit{w/o pre-training} & 83.01 & 91.23 \\
    \ \textit{w/ MTM+MVM} & 87.20 & 93.92 \\
    \ \textbf{\textit{w/ MTM+MVM+MGM}} & \textbf{89.44} & \textbf{93.64} \\
    \bottomrule
    \end{tabular}%
    }}
      \caption{The results of the ablation study for pre-training. Performance steadily increases as pre-training tasks are added.}
  \label{tab:ablation}%
\end{table}%

\begin{table}[t]
  \centering
  \small
  \resizebox{\linewidth}{!}{
    \setlength{\tabcolsep}{4mm}{
\begin{tabular}{lcc}
\toprule
\multirow{2}[2]{*}{\textbf{Model}} & \textbf{FUNSD} & \textbf{CORD} \\
      & \textbf{(F1)} & \textbf{(F1)} \\
\midrule
\MODEL \\
\ \textit{w/ Region}                 & 80.82 & 94.24 \\
\ \textit{w/ Region + Word}          & 86.96 & 95.49 \\
\ \textit{w/ Page + Region}          & 81.65 & 94.69 \\
\ \textbf{\textit{w/ Page + Region + Word}}   & \textbf{89.44} & \textbf{97.11} \\
\bottomrule
\end{tabular}%
    }}
      \caption{The results of the ablation study for multi-granular features. We observe a steady increment with more features involved, and the word-level features contribute most to the improvement.}
  \label{tab:ablation2}%
\end{table}%

To study the importance of the pre-training tasks, we design an ablation study that skips several pre-training tasks. The results are shown in Table \ref{tab:ablation}. In the first setting, we skip the entire pre-training stage so all the parameters can only be learned in the downstream tasks. In the second setting, we include the commonly-used masking techniques. The model is pre-trained with the two masking tasks in our design, the mask sentence modeling, and the mask vision modeling. Performance steadily increases as pre-training tasks are added; overall, pre-training improves the performance by 6.43\%, 2.69\% on FUNSD and RVL-CDIP, respectively. 

We believe that the masking strategy enables the model to learn from the multi-modal context of the page. In the third experiment in the table, we add the alignment techniques between words and regions designed to strengthen the connection between multiple granularities. The performance on FUNSD is further improved by 2.24\%, while there is also a decrease of 0.28\% in the performance on RVL-CDIP. Local connections between words and regions are helpful in fine-grained tasks but may introduce some noise to coarse-grained tasks.

To study the role of features from each granularity, we also conduct an ablation study using different combinations of multi-granular features, where we feed the model with features from region-level inputs, region-level and word-level inputs, page-level and region-level inputs, respectively. We report the performance on FUNSD and CORD in Table \ref{tab:ablation2}. We observe a steady increment with more features involved, and the word-level features contribute more to the improvement. 

\begin{figure*}[t]
    \centering
    \includegraphics[width=\linewidth]{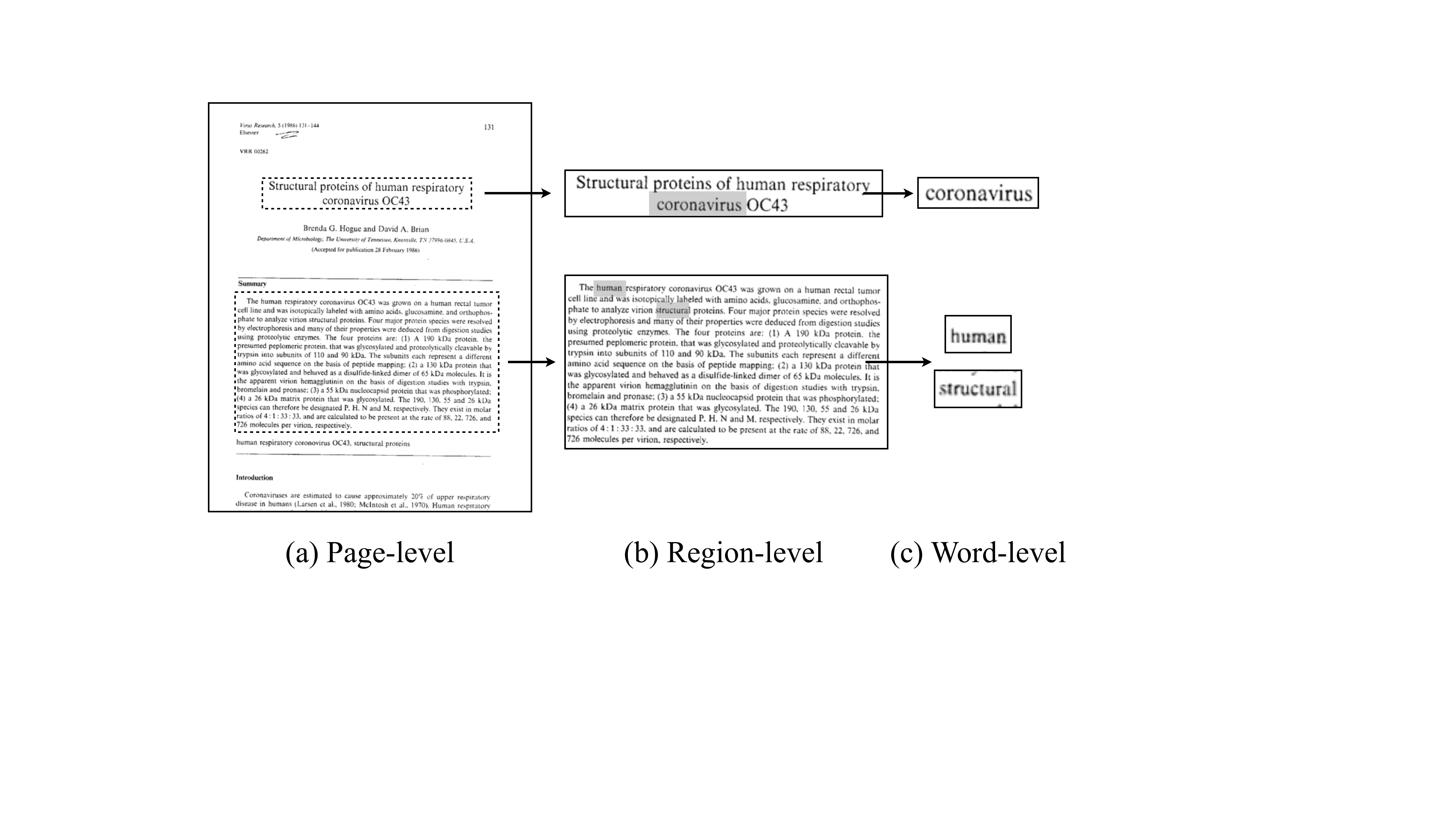}
    \caption{The error analysis and visualization}
    \label{fig:example}
\end{figure*}

\begin{figure}[t]
\centering
\subcaptionbox{\label{1}}{\includegraphics[width =.4\linewidth]{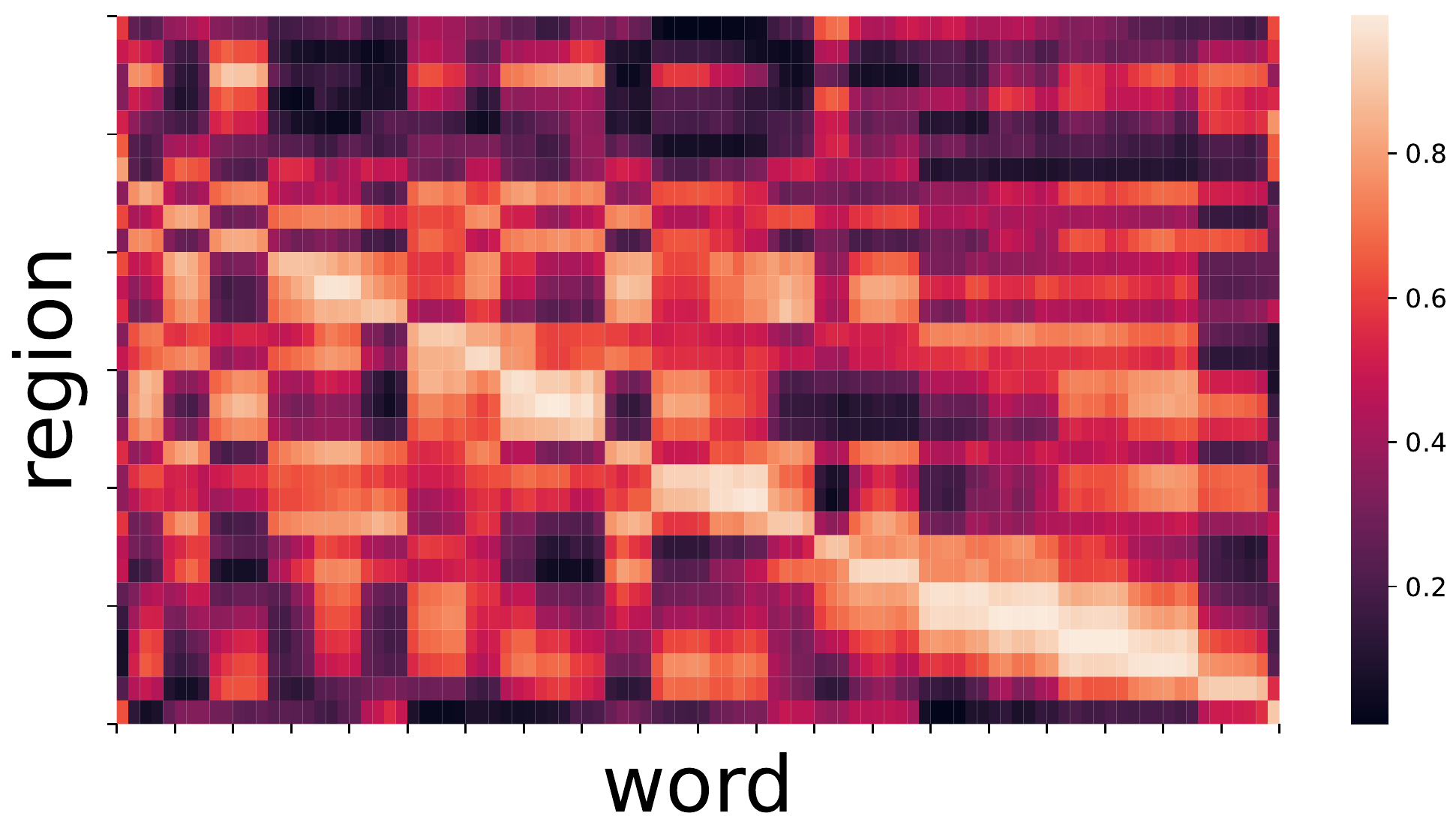}}
\subcaptionbox{\label{2}}{\includegraphics[width =.4\linewidth]{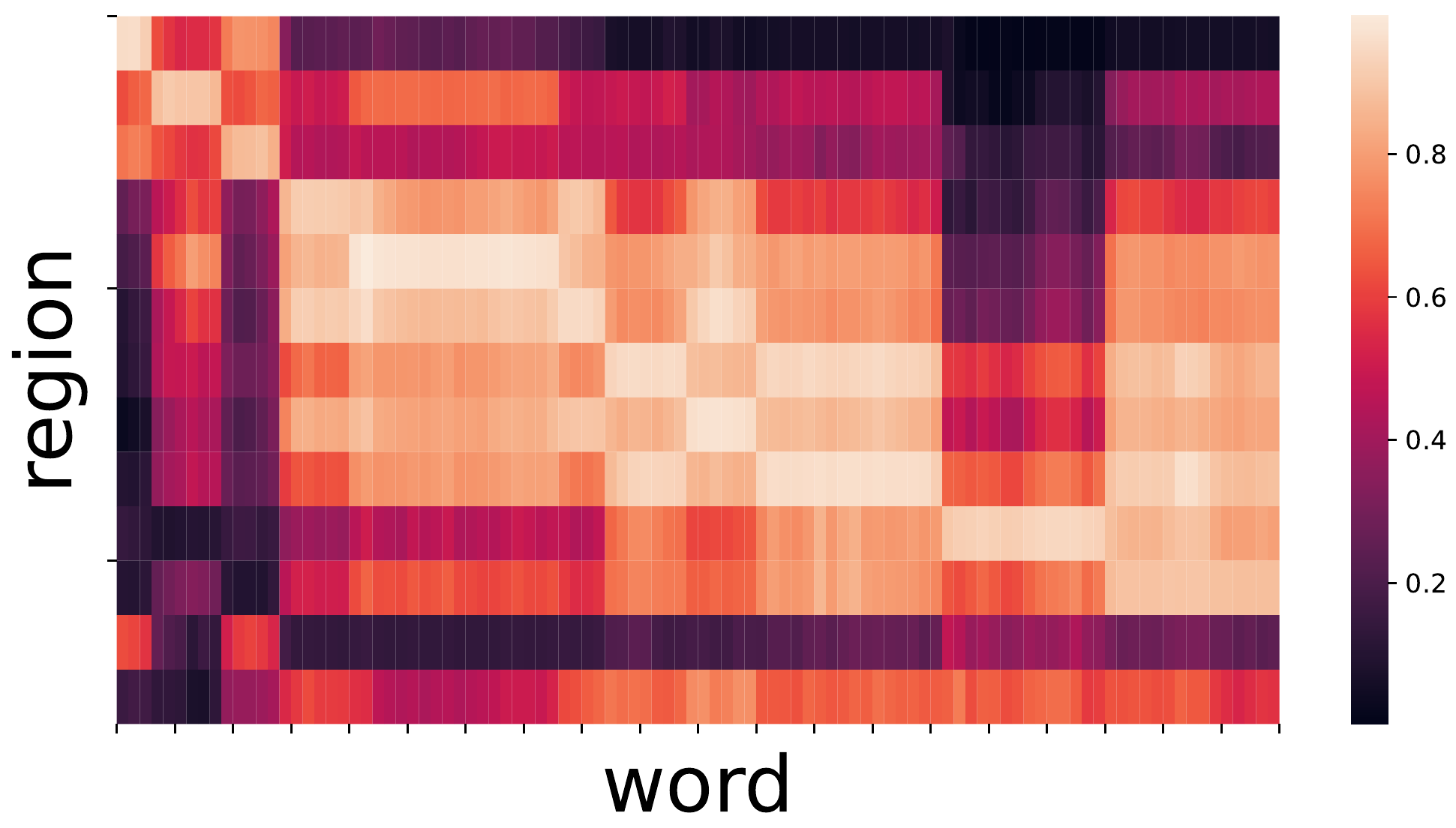}}
\\
\subcaptionbox{\label{4}}{\includegraphics[width =.4\linewidth]{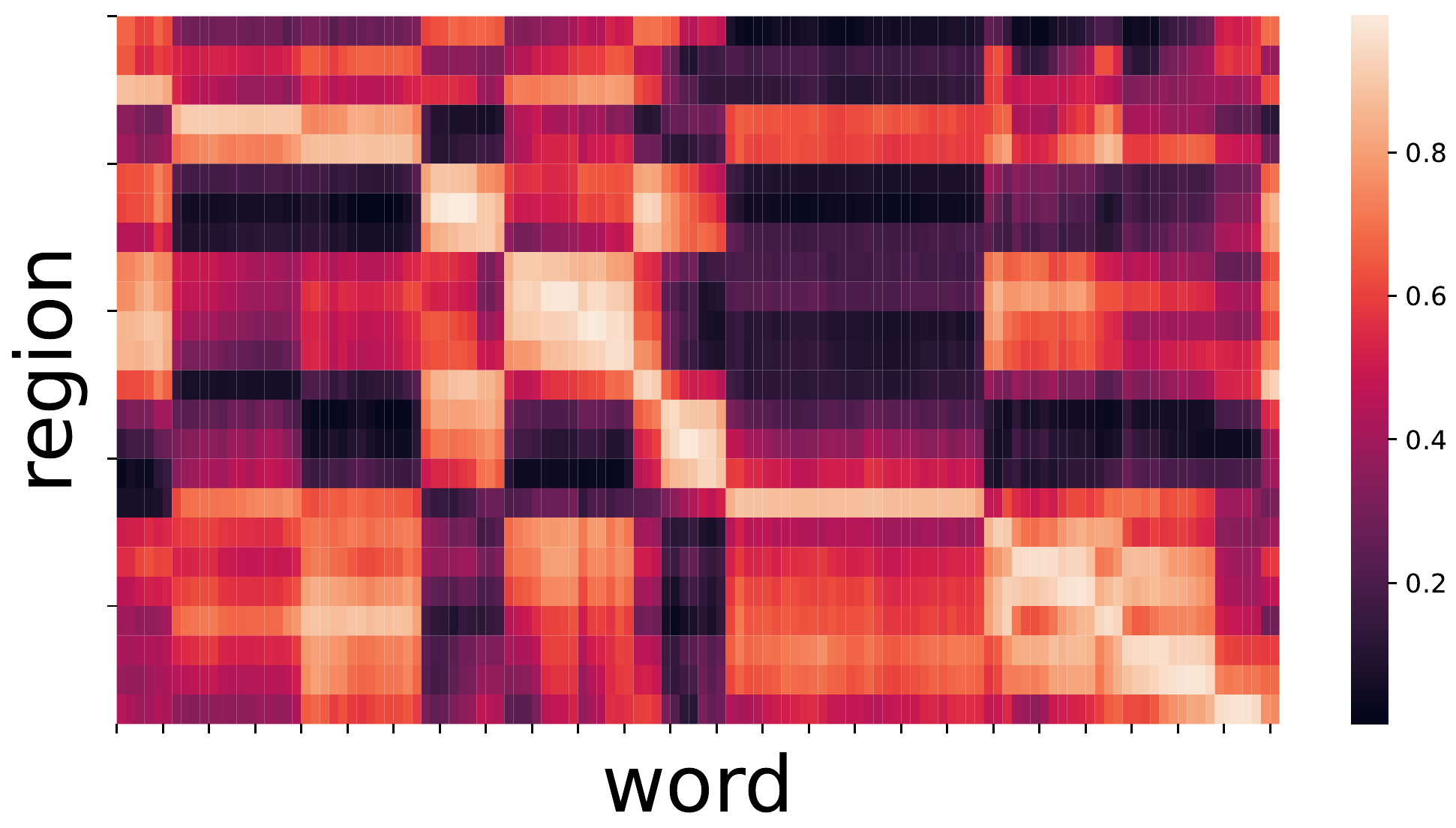}}
\subcaptionbox{\label{5}}{\includegraphics[width =.4\linewidth]{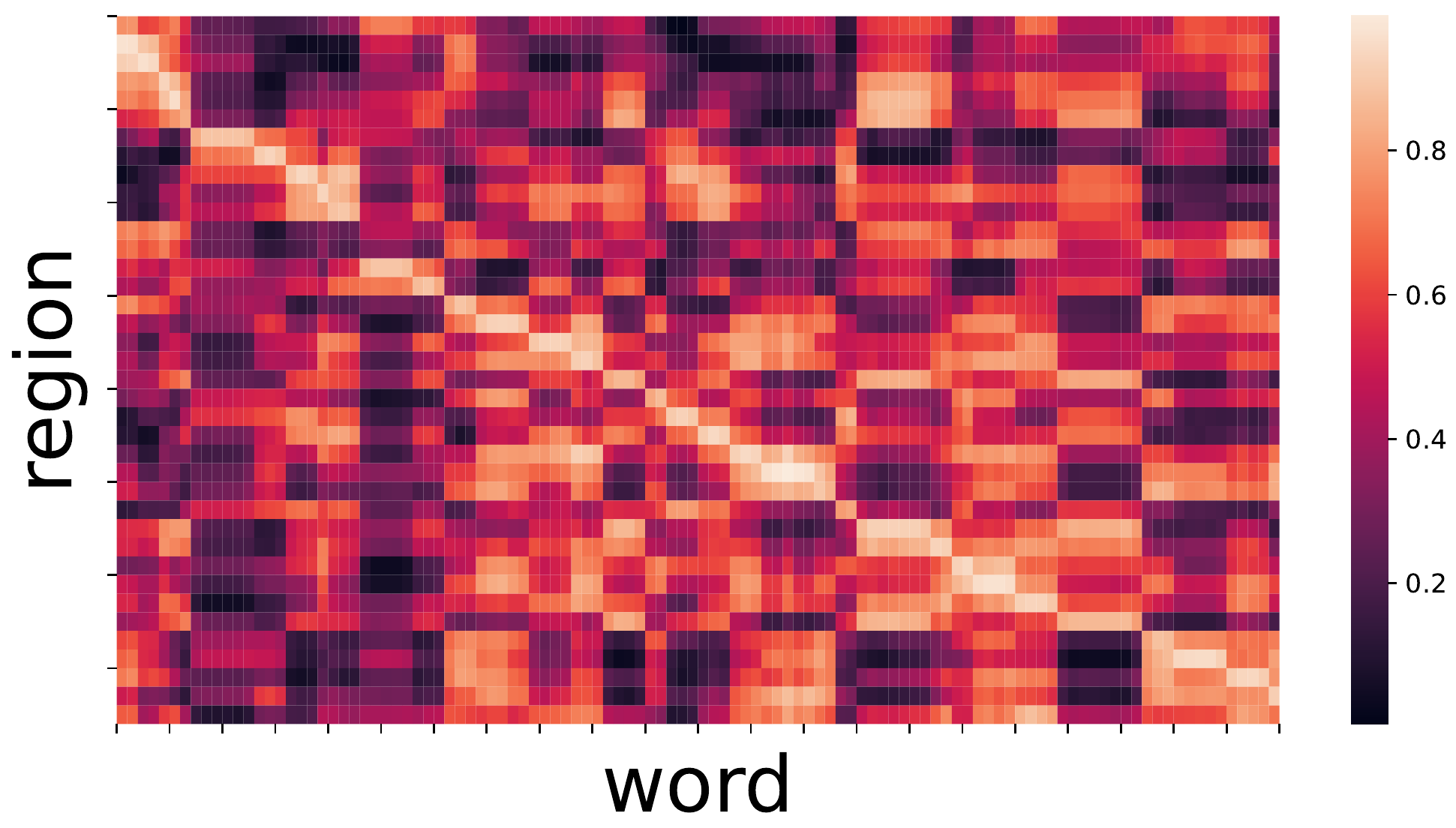}}
\caption{The correlation weight visualization between regions and words}
\label{fig:heatmap}
\end{figure}

\subsection{Region-word Correlation Visualization}\label{sec:alignemnt-visualization}

We visualize the correlation between regions and words using heat maps. We select four examples from the FUNSD dataset and show the heat maps of the final feature dot product in Figure \ref{fig:heatmap}. The x-axis and the y-axis correspond to the words and regions, respectively, and the lighter the color is, the higher correlation there is. Some cropping is applied for clearer visualization. From the heat maps, we can observe that there are highlighted areas along the matrix diagonal, which means our model learns the region-word hierarchy in the pre-training stage and can leverage such correspondence in downstream tasks. We also see some lighter colored blocks in the matrix. Since all the words and regions are serialized in positional order, these lighter colored blocks indicate the model is able to use the localized features in the model with the help of multi-granular inputs. This ability further confirms that our intuition that combining information from different levels of granularity will be beneficial is correct.

\subsection{Error Analysis}

We select several representative cases in the comparison between UDoc and \MODEL and show them in Figure \ref{fig:example}. We also visualize the weight matrix of the entities in the same way as in Section \ref{sec:alignemnt-visualization}. In these examples, our proposed model can leverage the more fine-grained signal from word-level inputs and make the correct prediction. In example 2, the entity is labeled as \textit{Answer} where the corresponding question, ``Fax No.:'', is at the top of this column. Due to the large distance of this question-answer pair, UDoc predicts the entity as \textit{Question}, while \MODEL can give the right prediction by directly learning from the digits inside of the text fields, which is a strong signal for answers. From the heat map, we can also see that a lighter color appears in the corresponding area of the entity. Meanwhile, the word-level information even strengthens the multi-modal features since it provides more details of a given text field. As we can observe in example 4, the entity ``File with:'' is likely to be \textit{Header} or \textit{Question} given its textual contents and location in the page, but \MODEL can predict from the rich visual features that this field is a part of normal text and less likely to be \textit{Header}; these rich inputs allow \MODEL to make the correct prediction where UDoc cannot. However, in example 3, both UDoc and \MODEL cannot predict correctly. The ground-truth label is \textit{Header} but both models predict the entity as \textit{Question}. The entity is not at the top of the page where the header entities are more likely to be located, so we attribute this error to the dependence of \MODEL to the spatial information. 


\section{Related Work}


\paragraph{Word-level Models}
Word-level models inherit the architecture of pure-text pre-trained language models. Word-level contextual information is encoded by a multi-layered transformer, and spatial and visual features are added to refine the representation. Inspired by the positional embeddings in \citet{vaswani2017attention,raffel2019exploring,dai2019transformer},  absolute or relative spatial features based on the bounding boxes are proposed to encode the words' layout with respect to each other~\cite{xu2020layoutlm,xu2020layoutlmv2,hong2021bros,garncarek2021lambert}.
Computer vision deep models~\cite{he2016deep,xie2017aggregated} are used to extract features from the document images, and self-supervised learning methods are applied to learn the cross-modal correlation between images and words~\cite{xu2020layoutlmv2,powalski2021going}.

\paragraph{Region-level Models}
Region-level models encode the regions in the document page including text blocks, headings, and paragraphs~\cite{li2021selfdoc,gu2021unidoc}. Similar spatial and visual features are used in these models as in the word-level models. With the help of coarse-grained inputs, region-level models can emphasize the rich locality features and catch high-level cues. Another difference with the word-level models is that the number of regions is much smaller than the word number on the page, so the region-level models are more efficient when processing long documents.



\section{Conclusions and Future Work}

We present \MODEL, a multi-modal multi-granular pre-training framework, which goes beyond the existing region-level or word-level models and leverages the contents at multiple levels of granularity to understand the document pages better. Existing models fail to use the informative multi-granular features in the document due to the restriction from the word-level model architecture, and lead to unsatisfactory results. We solve these issues with the new architecture design and tailored pre-training tasks. With a unified multi-modal encoder, we embed the features from pages, regions, and words into the same hyperspace, and design a multi-granular attention mechanism and multi-granularity modeling task for \MODEL to learn the spatial hierarchical relation between them. Experiments show that our proposed model can understand the spatial relation between the multi-granular features and lead to improvements in downstream tasks.

As for future work, since we have not fully exploited the multi-granular information, we will go beyond the page level and investigate the possibility of encoding multiple pages. We are also interested in inputs that are more fine-grained than word level, such as pixels.

\section{Acknowledgement}
This work was supported in part by Adobe Research. We thank anonymous reviewers and program chairs for their valuable and insightful feedback.
\section*{Limitations}
Although we inherit the idea of using region-level inputs from \citep{gu2021unidoc,li2021selfdoc}, we cannot keep their merits of saving computing resources. Region-level models encode regions instead of all the words in the page, so the smaller number of features are included in the self-attention layers. However, we want to leverage the fine-grained word-level information as \citep{xu2020layoutlm,xu2020layoutlmv2,hong2021bros}, so the words are also considered in the multi-granular attention and the multi-modal attention layers. Compared to existing works, our work requires more memory storage during training and testing.

\section*{Ethical Considerations}

This paper presents a new framework for document image understanding tasks. Our model is built on open-source tools and datasets, and we aim at increasing the efficiency of processing various documents and also bringing convenience to ordinary people's life. Thus, we do not anticipate any major ethical concerns. 

\bibliography{ref}
\bibliographystyle{acl_natbib}


\end{document}